\documentclass[narrow]{article}
\usepackage[utf8]{inputenc}
\usepackage{amsmath,amssymb,amsfonts}
\usepackage{array}
\usepackage[tableposition=top]{caption}
\usepackage{graphicx}
\usepackage{makecell}
\usepackage{stfloats}
\usepackage{authblk}
\usepackage{diagbox}
\usepackage[superscript]{cite}

\newcounter{allsec}

\newcommand{\words}[1]{}

\title{Benchmarking learned non-Cartesian k-space trajectories and reconstruction networks}

\author[1,2]{Chaithya G R}
\author[1,2]{Philippe Ciuciu}
\affil[1]{CEA, Joliot, NeuroSpin, Universit\'e Paris-Saclay, F-91191 Gif-sur-Yvette, France}
\affil[2]{Inria, Parietal, Universit\'e Paris-Saclay, F-91120 Palaiseau, France}

\renewcommand{\b}[1]{\mathbf{#1}}

\begin{document}
\date{}
\maketitle
\section*{Synopsis}
\refstepcounter{allsec}
We benchmark the current existing methods to jointly learn non-Cartesian k-space trajectory and reconstruction: PILOT\cite{pilot}, BJORK\cite{bjork} and compare them with those obtained from recently developed generalized {\em hybrid learning}~(HybLearn) framework\cite{hybrid_learning}.
We present the advantages of using projected gradient descent to enforce MR scanner hardware constraints as compared to using added penalties in the cost function.
Further, we use the novel HybLearn scheme to jointly learn and compare our results through retrospective study on fastMRI validation dataset.

	\words{\immediate\write18{texcount -sub=section main.tex  | grep "Section" | sed -e 's/+.*//' | sed -n \theallsec p > 'count.txt'}
	(\input{count.txt}words)}
\section*{Summary of main findings}
\refstepcounter{allsec}
Our method converges to improved k-space trajectories, having more accurate k-space coverage particularly at lower frequencies. It resulted in 3-4dB gain in PSNR and almost 0.06 gain in SSIM scores as compared to earlier state-of-the-art methods based on joint or alternated learning schemes.

	\words{\immediate\write18{texcount -char -sub=section main.tex  | grep "Section" | sed -e 's/+.*//' | sed -n \theallsec p > 'count.txt'}
	(\input{count.txt}characters)} 
\section{Introduction}
\refstepcounter{allsec}
Compressed Sensing in MRI involves the optimization of k-space sampling trajectories and image reconstruction from the undersampled k-space data.
In this regard, PILOT\cite{pilot, 3d-flat} was developed to learn k-space trajectory jointly with a U-Net as reconstruction network. 
However, it relies on auto-differentiation of the NUFFT operator, which may not be very accurate as shown in\cite{wang_ISMRM21}, thus resulting in suboptimal local minima.

More recently, BJORK\cite{bjork} learned trajectories using a more accurate Jacobian approximation of the NUFFT operator\cite{wang_ISMRM21} . 
Yet, both BJORK\cite{bjork} and PILOT\cite{pilot} enforce the hardware constraints as penalty terms in the overall loss function, thus requiring the tuning of at least one hyperparameter associated with these terms. 
Additionally, such penalty could affect the overall gradients, thereby resulting in suboptimality. 
Further, BJORK\cite{bjork} was parameterized with B-spline curves, which could severely limit the shape of trajectories and prevent them from better exploring the k-space. 
Finally, both above mentioned methods do not make use of any density compensation~(DCp) mechanism for image reconstruction, although the latter plays a critical role in obtaining cleaner MR images 
in the non-Cartesian deep learning setting\cite{ncpdnet}.

In this work, we compare BJORK\cite{bjork} and PILOT\cite{pilot} with the proposed generic {\em hybrid} framework\cite{hybrid_learning} for learning k-space trajectories with projected gradient descent.

	\words{\immediate\write18{texcount -sub=section main.tex  | grep "Section" | sed -e 's/+.*//' | sed -n \theallsec p > 'count.txt'}
	(\input{count.txt}words)}
\section{Model and Notation}
\begin{figure}[h!]
	\includegraphics[width=\textwidth]{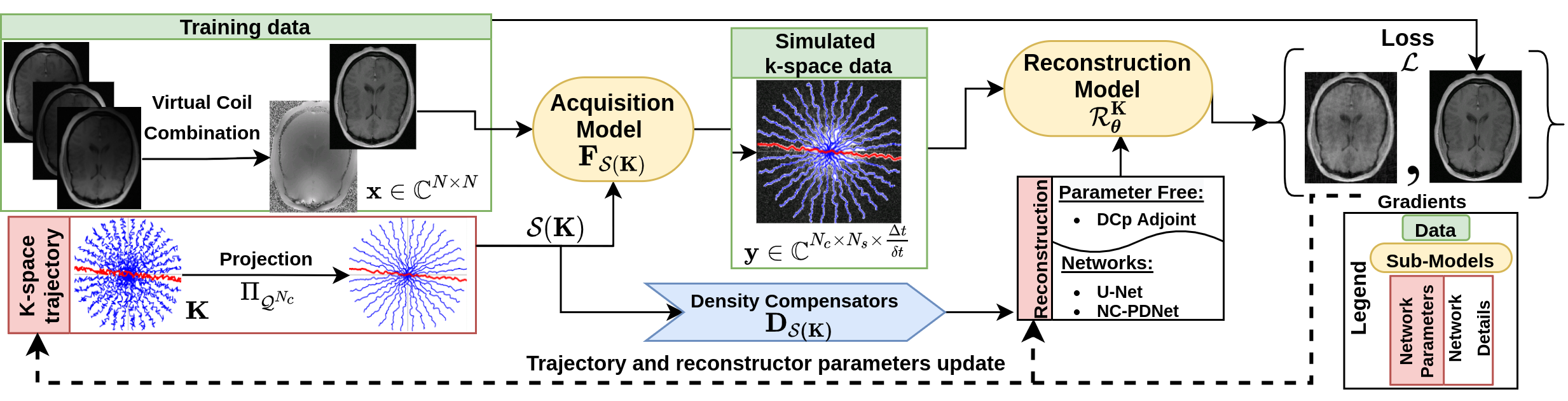}
	\caption{\label{fig:gen_model} A generic learning-based framework for joint optimization of the MRI acquisition and reconstruction models. }
\end{figure}

\refstepcounter{allsec}
We used the generic model (Fig.\ref{fig:gen_model}) developed in\cite{hybrid_learning} to learn hardware compliant k-space trajectories $\b{K}$.
The model was trained on complex-valued brain images $\b{x}$ obtained by virtual coil combination\cite{parker2014phase} of the per-channel images in fastMRI  dataset\cite{zbontar2018fastmri}, to account for the phase accrual and make the forward model more realistic. 
Here, projection $\Pi_{\mathcal{Q}_{N_c}}$ was carried out after every gradient descent step to enforce constraints.
Later on, the trajectories are interpolated with a linear operator $\mathcal{S}$ to model the analog-to-digital converter in the scanner. 
The acquisition model is simulated by a forward NUFFT operator $\b{F}_{\mathcal{S}(\b{K})}$. 
The density compensator $\b{D}_{\mathcal{S}(\b{K})}$ is estimated with\cite{pipe_dc} and is used by reconstruction network $\mathcal{R}_{\boldsymbol{\theta}}^\b{K}$ giving reconstructed image $\widehat{\b{x}}$.

	\words{\immediate\write18{texcount -sub=section main.tex  | grep "Section" | sed -e 's/+.*//' | sed -n \theallsec p > 'count.txt'}
	(\input{count.txt}words)}
\section{Methods}
\refstepcounter{allsec}
The model described above was trained with combined L1-L2-MSSIM loss $\mathcal{L}$ as described in\cite{hybrid_learning}.
The trajectories were learned with ADAM optimizer and reconstruction network $\mathcal{R}_{\boldsymbol{\theta}}^\b{K}$ was trained with Rectified-ADAM.
The training was done with a learning rate of $10^{-3}$ and batch size of 64 on the fastMRI training data, which was split into training and validation in a 90\%-10\% ratio.
This enabled early stopping to prevent overfitting.
The original fastMRI validation dataset was used only for later evaluation.
The entire training was carried out at different resolution levels and using HybLearn as presented in Sec.3.3 and Tab.1 in\cite{hybrid_learning}.

We learned k-space trajectories with $N_c=16$ shots and $N_s=512$ samples per shot (observation time $T_{\textrm{obs}}=5.12\textrm{ms}$, raster time $\Delta\textrm{}t=10\text{µs}$, dwell time $\delta\textrm{}t=\text{2µs}$).
For comparison with an earlier baseline, we use SPARKLING trajectories generated with the learned sampling density using LOUPE\cite{loupe2020} as obtained in\cite{learning_density} and trained NC-PDNet\cite{ncpdnet} as a reconstructor for it.

We compared our results with PILOT and BJORK trajectories, which were obtained directly from the respective authors. As we didn't receive their trained reconstruction networks, we trained NC-PDNet by ourselves for a fair comparison: NC-PDNet makes use of DCp and its Cartesian version stood 2nd in the 2020 fastMRI challenge\cite{fastmri_challenge}. This way, we used the same reconstructor for all the trajectories, with the same network parameters and which was trained individually. Our comparison with PILOT (Fig.\ref{fig:pilot}) was carried out for $T_1$ and $T_2$ contrasts in the fastMRI dataset.

As the BJORK trajectory was learned for $\Delta\textrm{}t=4\textrm{µs}$, to ensure fair comparison, we obtained trajectories with the same specifications. This comparison (Fig.\ref{fig:bjork}) was done at different undersampling factors~(UF).

	\words{\immediate\write18{texcount -sub=section main.tex  | grep "Section" | sed -e 's/+.*//' | sed -n \theallsec p > 'count.txt'}
	(\input{count.txt}words)}

\section{Results}
\begin{figure}[h!]
	\includegraphics[width=\textwidth]{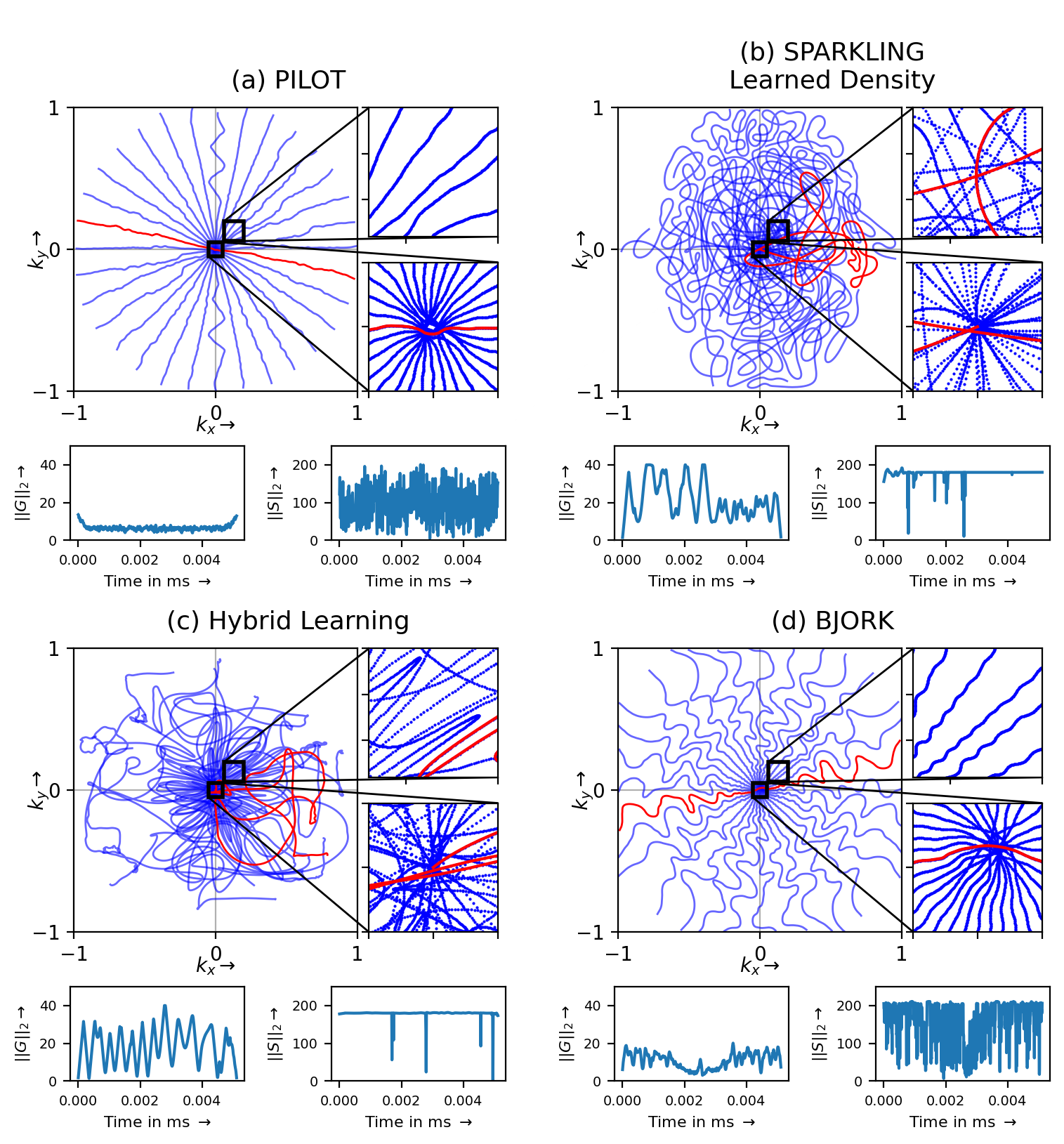}
	\caption{The optimized hardware compliant non-Cartesian k-space trajectories using \textbf{(a)} PILOT, \textbf{(b)} SPARKLING with learned density using LOUPE, \textbf{(c)} HybLearn scheme, \textbf{(d)} BJORK. The number of shots $N_c$=16. The number of dwell time samples are set to match the same number of sampling points overall. 
	Zoomed in visualizations of the center of k-space (bottom) and slightly off-center (top) is presented at the right of corresponding trajectories.
	The corresponding gradient $||G||_2$ (in mT/m) and slew rate $||S||_2$ (in T/m/s) profiles are depicted below each trajectory.}
	\label{fig:traj}
\end{figure}

\refstepcounter{allsec}
When comparing the zoomed portions of optimized trajectories (Fig.\ref{fig:traj}), we observe that PILOT has a k-space hole at the center while BJORK samples the k-space densely slightly off the center, which is suboptimal. 
In contrast, HybLearn and SPARKLING methods sample the central region of k-space more densely, which could help obtain improved image quality. 

We see that PILOT and BJORK do not efficiently use the gradient hardware and have similar gradient and slew rate profiles, while SPARKLING and HybLearn trajectories, are hitting the gradient constraints more often for the maximal gradient and almost everywhere for the slew rate. 
This difference could be attributed to using a projector for hardware constraints as compared to handling a penalty.

Next, we compared the retrospective results with PILOT (Fig.\ref{fig:pilot}) and BJORK (Fig.\ref{fig:bjork}) obtained on 512 slices from fastMRI validation dataset. We observe that both SPARKLING with a learned density and HybLearn outperform PILOT and BJORK, with HybLearn performing the best with a gain of nearly 0.06 in SSIM and 3-4dB in PSNR scores as compared to PILOT and BJORK.

\newcommand{\filenametone}{file\_brain\_AXT1PRE\_209\_6001221.h5}
\newcommand{\filenamettwo}{file\_brain\_AXT2\_205\_2050175.h5}
\begin{figure}[h!]
	\includegraphics[width=\textwidth]{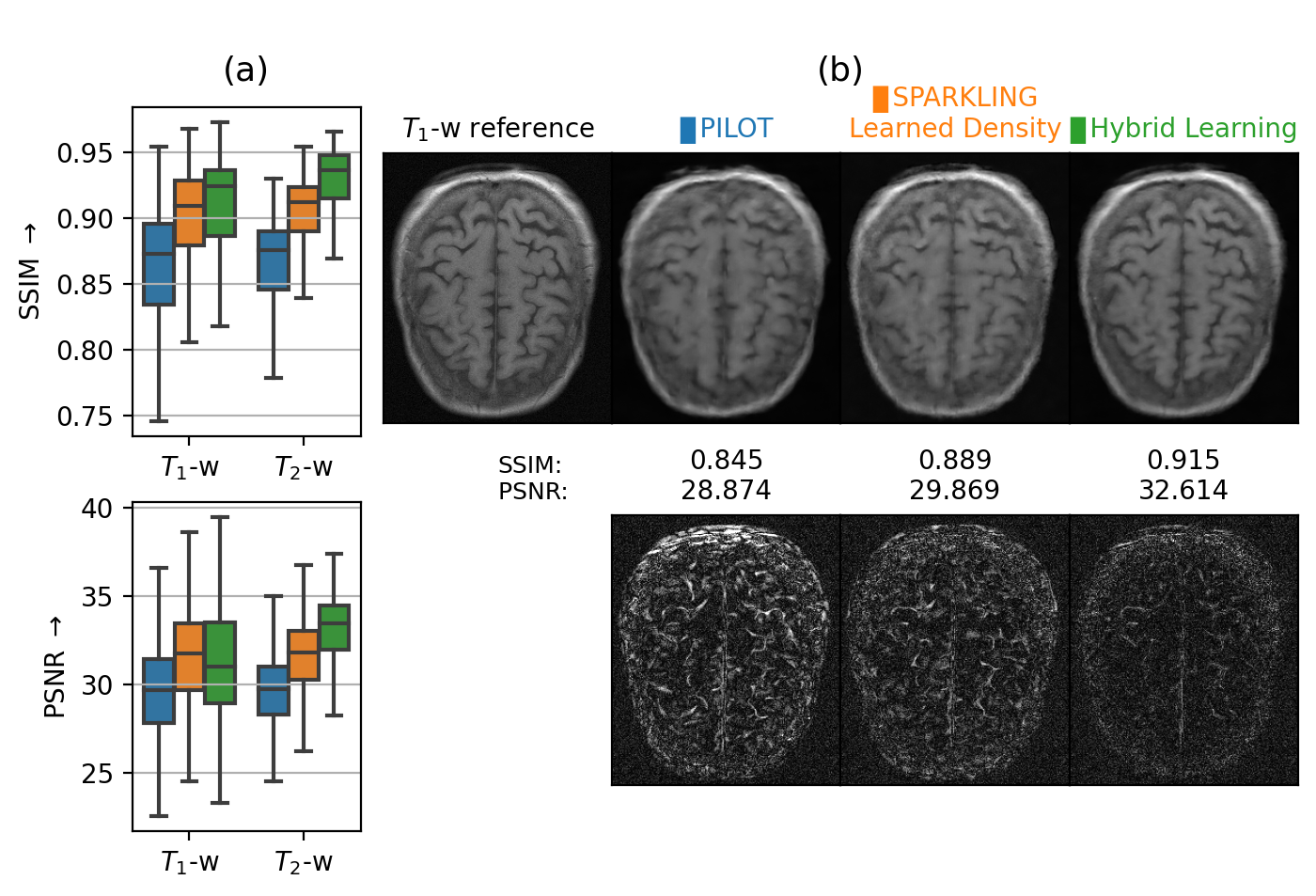}
	\caption{\textbf{(a)} Box plots comparing the image reconstruction results on a retrospective study at UF=2.5 ($N_c=16$, $N_s=512$, $\frac{\Delta t}{\delta t}=5$) using 512 slices of $T_1$ and $T_2$ contrasts (fastMRI validation dataset) using PILOT (blue), SPARKLING with learned density (orange) and HybLearn (green). SSIMs/PSNRs appear on top/at the bottom. \textbf{(b) Top:} $T_1$-w reference image and reconstruction results for a single slice from \texttt{\filenametone} with corresponding strategies. \textbf{(b) Bottom:} The residuals, scaled to match and compare across methods.}
	\label{fig:pilot}
\end{figure}

\begin{figure}[h!]
	\includegraphics[width=\textwidth]{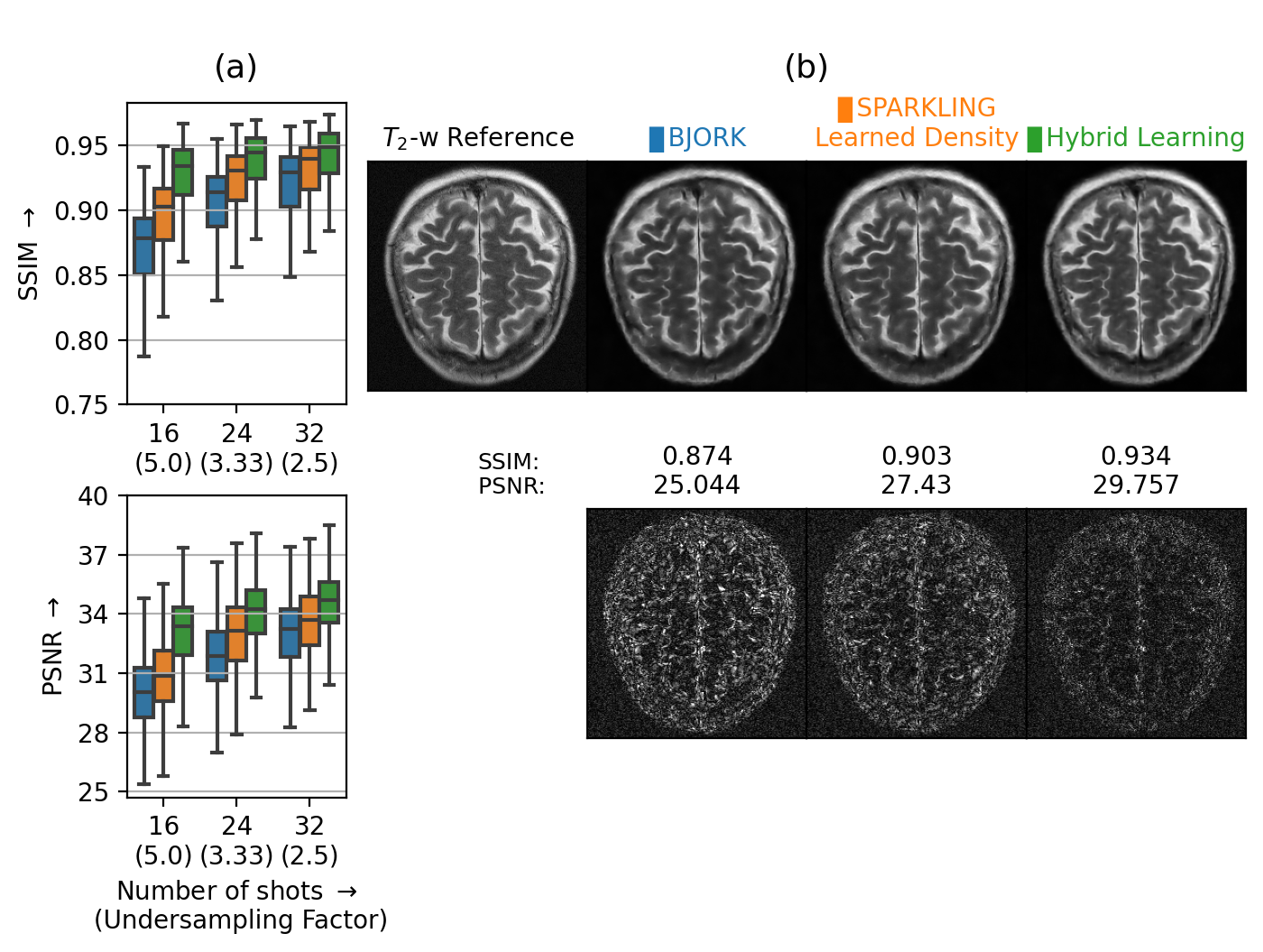}
	\caption{\textbf{(a)} Box plots comparing the image reconstruction results on a retrospective study using 512 slices on $T_2$ contrast (fastMRI validation dataset) using BJORK (blue), SPARKLING with learned density (orange) and HybLearn (green). We present the results at varying undersampling factors characterized with $N_c=16$, 24 and 32. SSIMs/PSNRs appear on top/at the bottom. \textbf{(b) Top:} $T_2$-w reference image and reconstruction results for a single slice from \texttt{\filenamettwo} with corresponding strategies. \textbf{(b) Bottom:} The residuals, scaled to match and compare across methods.}
	\label{fig:bjork}
\end{figure}

	\words{\immediate\write18{texcount -sub=section main.tex  | grep "Section" | sed -e 's/+.*//' | sed -n \theallsec p > 'count.txt'}
	(\input{count.txt}words)}

\section{Conclusion}
\refstepcounter{allsec}
In this work, we benchmarked the trajectories obtained by HybLearn with PILOT\cite{pilot} and BJORK\cite{bjork}.
Although the learned neural networks in PILOT and BJORK were not available for a full end-to-end comparison, we performed a fair assessment by training a NC-PDNet\cite{ncpdnet} as common reference for image reconstruction. 
Through restrospective studies on the fastMRI validation dataset, we showed that this hybrid learning scheme works across multiple resolutions and leads to superior performance of the trajectories and improved image quality overall.

Future prospects of this work include prospective implementations through modifications of $T_1$ and $T_2$-w imaging sequences.

	\words{\immediate\write18{texcount -sub=section main.tex  | grep "Section" | sed -e 's/+.*//' | sed -n \theallsec p > 'count.txt'}
	(\input{count.txt}words)}

	\words{\immediate\write18{texcount -sub=section main.tex  | grep "Section" | sed -e 's/+.*//' | tail -n+3 | head -n-1| awk '{ sum += $1 } END { print sum }' > 'total.txt'}
	Total Words =  \input{total.txt}words}

\section*{Acknowledgements}
\refstepcounter{allsec}
This work was granted access to the HPC resources of IDRIS under the allocation 2021-AD011011153 made by GENCI. Chaithya G R was supported by the CEA NUMERICS program, which has received funding from the European Union's Horizon 2020 research and innovation program under the Marie Sklodowska-Curie grant agreement No 800945. We would like to thank the authors of BJORK\cite{bjork} and PILOT\cite{pilot} for sharing the trajectories for comparisons done in this abstract.

\bibliographystyle{unsrt}
\bibliography{ref}
\end{document}